\documentclass[a4paper,oneside,11pt]{article} 

\usepackage[latin1]{inputenc}
\usepackage[T1]{fontenc}     
\usepackage{amsmath, amsthm} 
\usepackage{amsfonts,amssymb}

\usepackage{graphicx}        
\usepackage{subfig}          
\usepackage{wrapfig} 

\usepackage{array}           
\usepackage{multirow}        

\usepackage{listings}        
\usepackage{color}           
\usepackage{fancybox}        
\usepackage{bbm}             
\usepackage{dsfont}          
\usepackage{fancyhdr}        
\usepackage{natbib}

\usepackage{hyperref}
 \hypersetup{colorlinks, linkcolor=black, urlcolor=black} 

\makeatletter
\newif\if@restonecol
\makeatother

\usepackage{algorithm2e}
\usepackage{multirow}

\newcommand{\OMIT}[1]{}
\newcommand{\vb}{\mathbf{v}}    
\newcommand{\lonetwo}{\ell_1/\ell_2}

\newcommand{\s}[1]{\text{supp}\left(#1\right)}

\newcommand{\VV}{\mathbb V}
\newcommand{\PP}{\mathbb P}

\newcommand{\RR}{\mathbb{R}}

\newcommand{\G}{\mathcal{G}}
\newcommand{\Ogroup}[1]{\Omega_{\text{group}}^{\G}\left(#1\right)}
\newcommand{\Om}[1]{\Omega_{\text{overlap}}^{\G}\left(#1\right)}
\newcommand {\nm}[1]{\left\|#1\right\|}

\begin{document}

\title{Improving stability and interpretability of gene expression signatures}

\author{
\small\bf Anne-Claire Haury\thanks{To whom correspondance should
    be addressed: \small 35, rue Saint Honor\'e, F-77300 Fontainebleau, France.}\\
\small Mines ParisTech, CBIO\\
\small Institut Curie, Paris, F-75248\\
\small INSERM, U900, Paris, F-75248\\
\small \texttt{anne-claire.haury@ensmp.fr}
\and
\small\bf Laurent Jacob\\
\small Mines ParisTech, CBIO\\
\small Institut Curie, Paris, F-75248\\
\small INSERM, U900, Paris, F-75248\\
\small \texttt{laurent.jacob@ensmp.fr} \\
\and
\small\bf Jean-Philippe Vert \\
\small Mines ParisTech, CBIO\\
\small Institut Curie, Paris, F-75248\\
\small INSERM, U900, Paris, F-75248\\
\small \texttt{jean-philippe.vert@mines-paristech.fr} \\
}


\maketitle

\begin{abstract}
\strut\newline
  \begin{description}
  \item[Motivation] Molecular signatures for diagnosis or prognosis
    estimated from large-scale gene expression data often lack
    robustness and stability, rendering their biological
    interpretation challenging. Increasing the signature's
    interpretability and stability across perturbations of a given
    dataset and, if possible, across datasets, is urgently needed to
    ease the discovery of important biological processes and,
    eventually, new drug targets.

    \OMIT{We address the widely investigated problem of finding a
      signature to predict breast cancer outcome from tumoral gene
      expression data. While many signatures have already been
      discovered that predict the metastatic status of the patients
      with some success, it remains that they have very few genes in
      common, if any. Moreover, studies have shown that many different
      sets of genes could indeed be used to obtain an acceptable
      predictive performance. Our concern is therefore to find a way
      to improve the signature's stability, that is, to derive a list
      of genes robustely selected across perturbations of a dataset
      and, if possible, across different datasets.  }
  \item[Results] We propose a new method to construct signatures
    with increased stability and easier interpretability. The method
    uses a gene network as side interpretation and enforces a large
    connectivity among the genes in the signature, leading to
    signatures typically made of genes clustered in a few
    subnetworks. It combines the recently proposed graph Lasso
    procedure with a stability selection procedure. We evaluate its
    relevance for the estimation of a prognostic signature in breast
    cancer, and highlight in particular the increase in
    interpretability and stability of the signature.

    \OMIT{ One class of algorithms that successfully perform variable
      selection under some conditions are the $L1$ penalized
      classifiers, such as theLasso. From that same class is the Group
      Lasso algorithm that allows the sparse selection of sets of
      variables. However neither the Lasso nor the Group Lasso cope
      well with too correlated features. We use randomization
      techniques to help overcome this issue and thus stabilize the
      signature.

      We propose to choose the groups from a protein-protein
      interaction graph and experimentally show that both this way of
      grouping the genes and randomization help improve the stability
      and the relevancy of the signature.  }
  \item[Availability] The code and data are available upon request.
  \item[Contact] \href{anne-claire.haury@mines-paristech.fr}{anne-claire.haury@mines-paristech.fr}
  \end{description}
\end{abstract}

\section{Introduction}
In recent years a large number of diagnostic, prognostic and predictive molecular signatures have been identified through analysis of genome-wide expression profiles \citep{Golub1999Molecular,Alizadeh2000Distinct,Ramaswamy2001Multiclass,Vijver2002gene-expression}. Common signatures involve a few tens of genes whose expression levels allow to classify a sample in a given disease subtype, or assess its prognosis. They have been quickly adopted by the medical community for their ability to provide accurate classification and prediction, and for their direct usefulness in the clinical context. For example, the 70-gene MammaPrint signature is now marketed as a molecular diagnostic test to assess the risk of metastasis for breast cancer \citep{Vijver2002gene-expression}.

Besides their predictive accuracy, signatures should bring useful biological information for further biomedical research, such as the identification of genes or pathways with strong prognostic power which may lead to a new understanding of the underlying biology, and eventually to the identification of new drug targets. However, the signatures proposed in different studies have generally very few genes in common, and it is now well documented that many non-overlapping signatures can have similar predictive accuracy \citep{ein2005outcome}. The lack of stability of signatures across datasets can also be observed within a given dataset, as signatures obtained after random perturbations of a given dataset can also have poor overlaps, i.e., lack stability \citep{Abeel2009Robust}. An unfortunate consequence of this lack of stability is that the biological interpretation of possible functions and pathways underlying the signature is difficult \emph{a posteriori}.

To remedy the lack of stability and the difficult interpretation of signatures, several authors have proposed to use side information, such as known biological pathways and gene networks, to analyze expression data and build signatures. For example, \citet{Chuang2007Network-based} identifies groups of connected genes in the network (subgraphs) differentially expressed between two conditions; \citet{Rapaport2007Classification} proposed a formulation of support vector machines (SVM) to estimate a predictive model by constraining the weights of connected genes to be similar, allowing to associate positive or negative contributions to regions of the network. These approaches assume that connected genes should contribute similarly to the class prediction, by computing average expression over subnetworks or assuming similar predictive weights of connected genes; however one may argue that this is too strong an hypothesis for many networks.

Here we investigate a related question: how to estimate a molecular signature, typically of a few tens of genes, that would be "coherent" with a given gene network given \emph{a priori} in the sense that genes in the signature would tend to be connected to each other in the network. Note that here we do not want to constrain connected genes in the signature to have similar weights, we would just like them to be clustered in a limited number of connected components of the graph. The resulting connected components could then be more amenable to biological interpretation than individual genes, and could potentially be more stable across datasets due to the soft constraint induced on the choice of genes.

We assess the relevance of a new method named the \emph{graph Lasso}, proposed recently by \citet{jacob2009glo}, to automatically learn such a signature given a training set of expression data and a gene network. The graph Lasso is an extension of the Lasso regression \cite{Tibshirani1996Regression}, a widely-used and state-of-the-art method for feature selection and identification of sparse signature. In graph Lasso, the penalty used in the Lasso is modified to incorporate the gene network information, leading to the selection of features that are often connected to each other. The resulting algorithm is a convex optimization problem, whose unique solution can be found by efficient optimization methods. While the graph Lasso increases the interpretability of the signature by increasing the number of network edges between its components, it may suffer from lack of stability like many other feature selection methods including its cousin the Lasso. Recently randomization and aggregation have been proposed as a powerful way to increase the stability of feature selection methods in large dimension \cite{Abeel2009Robust}. To further increase the stability of the graph Lasso, we propose a procedure akin to stability selection \citep{meinshausen2008stability} in this context.

We evaluate the relevance of the resulting procedure for the estimation of a prognostic signature in breast cancer. We highlight in particular the increase in interpretability and stability resulting from the incorporation of a large gene network in the graph Lasso procedure, coupled with stability selection.

\section{Methods}

\subsection{Learning a signature with the Lasso}
Given a training set of gene expression data for $p$ genes in $n$ samples belonging to two classes (e.g., good and poor prognosis tumor samples), estimating a discriminative signature is a typical problem of \emph{feature selection} for supervised classification. For example, a popular approach in bioinformatics is to select genes by ranking them according to their correlation with the class information \citep[eg.,][]{Veer2002Gene}. Once genes are selected, it is necessary to estimate a predictive model using these genes only. In this
study, we build on a different and increasingly popular approach in statistical
learning where the selection of features and the estimation of a predictive model using this features are more tightly coupled. For example, one may look for a model which predicts the outcome as
well as possible under the constraint of involving as few genes as
possible. A direct formulation of this joint requirement is~:
\begin{equation}
  \label{eq:fs}
  \beta^{sig}=\arg\min_{\beta\in\mathbb{R}^p} L(\beta,\text{data}) + \lambda
  \sum_{j=1}^p\mathbf{1}_{\{\beta_j \neq 0\}},
\end{equation}
where $L$ is a function measuring the error made by using
$\beta$ to predict the outcome on the data, and $\mathbf{1}_{\{\beta_j
  \neq 0\}}$ is $1$ if parameter $\beta_j$ is non-zero, $0$ otherwise,
so that the second term counts the number of non-zero elements in
$\beta$. If $\beta$ contains few zeros, many genes can be
involved in the prediction and it is easy to make few errors
on the training data, corresponding to small values for $L$.  Conversely, if $\beta$ is very sparse, then is becomes more difficult to discriminate the training set correctly. The optimum $\beta^{sig}$ is a trade-off between
these two extremes. The hyperparameter $\lambda\geq
0$, which must be fixed before optimization, adjusts this
tradeoff~: at one extreme ($\lambda=0$) all the genes are involved in
the model, and at the other extreme we obtain $\beta^{sig} = 0$ (no
gene involved). Now the exact solution of problem~\eqref{eq:fs} cannot
be computed even for a reasonable number of genes, due to the combinatorial nature of the problem. This motivates the
introduction of methods such as the
Lasso~\citep{Tibshirani1996Regression}, where the second term is
replaced by $\|\beta\|_1 \stackrel{\Delta}{=} \sum_{j=1}^p|\beta_j|$. The new problem can be solved exactly, and also results in efficient feature selection.

\subsection{The Graph Lasso}

The group Lasso~\citep{Yuan2006Model} is a useful variant of the Lasso when the features are clustered into groups a priori, and one wishes to select features \emph{by groups}. It replaces the $\|\beta\|_1$ term in the Lasso formulation by
$\sum_{g\in\G}\|\beta_g\|$, where $\G$ is the set of groups of
variables which should be either all zero or all non-zero. Like
$\|\beta\|_1$ approximates the behavior of the count of selected
genes, $\sum_{g\in\G}\|\beta_g\|$ approximates the count of groups
which have at least one non-zero gene, and leads to solutions where
several groups contain only genes at $0$, which is exactly equivalent
to selecting groups in $\G$ as long as $\G$ is a \emph{partition} of
the genes, \emph{i.e.}, that each gene belongs to one and only one
group.

When some genes belong to several groups, a situation which arises for
example when considering gene pathways as groups, the group lasso does
not result anymore in the selection of a union of groups.
In~\cite{jacob2009glo}, a generalized version of this penalty was
proposed which allows to select unions of pre-defined groups which potentially overlap, \emph{e.g.} the pathways. The overlapping group
lasso penalty was empirically shown to select fewer groups than the
simple Lasso, and some results were given on its statistical
properties, in particular its model selection consistency.

Another interesting case which can be handled by this last penalty is
when a graph is defined on the genes, for example to represent
biological information such as co-regulation or protein-protein
interaction. In this case, finding a signature which is formed by few
connected subgraphs instead of a mere list of genes can make the
solution more interpretable as it defines new gene sets which are
optimal to predict the outcome~\citep{Chuang2007Network-based}. To
obtain this effect, one can simply use an overlapping group lasso
penalty, and define the groups to be the edges of the graph. Since the overlapping group
lasso leads to solutions in which a union of groups is selected, and since a 
union of is more likely to form few connected
subgraphs than randomly chosen genes, one can expect that the solution
will tend to form connected components. This effect was observed on
some simple examples in~\cite{jacob2009glo}. Here we investigate this
effect more thoroughly on an outcome prediction problem.

\subsection{Stability selection}

An issue with many feature selection methods, including the Lasso, is their lack of stability in the presence of many highly correlated features, which is to be expected with gene expression. In order to improve stability of feature selection, randomization and aggregation have been proposed as a powerful way to increase the stability of feature selection methods in large dimension \citep{meinshausen2008stability,Abeel2009Robust}. The general idea is to repeat the feature selection process on many randomly perturbed training sets (e.g., by bootstrapping the samples in the original training set), and to keep the features that are often selected in this procedure.

We propose a \emph{group selection} procedure to the graph lasso algorithm based on \citep{meinshausen2008stability}. The baseline of this procedure is shown in algorithm \ref{algo:stability}. 
\begin{algorithm}
\KwIn{Data $Z=(X,Y)$ divided into a training and a test sets, number of draws $ndraw$, $\Lambda$ a grid}
\KwOut{Probabilities  $(\Pi_j^{\lambda})_{g=1...pgroups, \lambda\in \Lambda}$}
\For{$i\in\{1...ndraw\}$}
{Draw $I$ a subsample of $\{1...n\}$ of size $[n/2]$ without replacement\;
\For{$\lambda \in \Lambda$}
{
Run a variable selection algorithm on $I$ with regularization parameter $\lambda$\;
Store the active set $\mathcal{A}(I,\lambda)$\;  
}
}
\For{$g\in\{1...pgroups\}$}
{
\For{$\lambda\in\Lambda$}
{Compute the \emph{selection probability} $\Pi_g^{\lambda}=\PP(g\in\mathcal{A}(I,\lambda)|I)$\;
} 
}
\caption{Stability Selection}
\label{algo:stability}
\end{algorithm}

\begin{figure}[!ht]
\centering
\includegraphics[width=\textwidth]{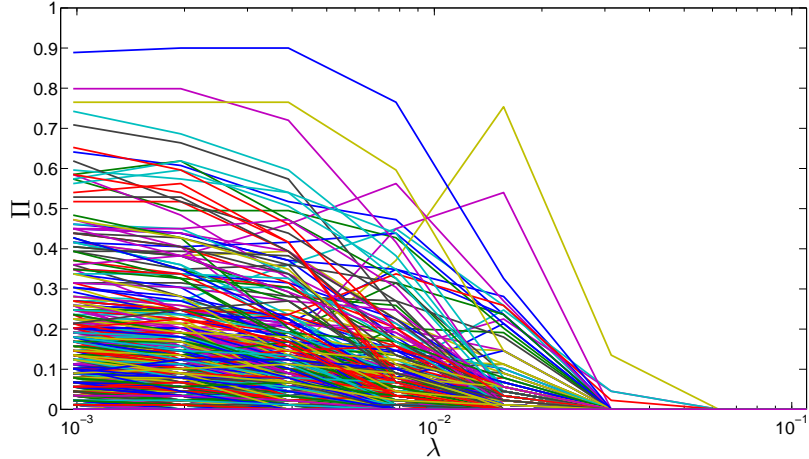}
\caption{Stability selection scores for all edges, as a function of $\lambda$.}
\label{fig:stab}
\end{figure}

\begin{figure}[!ht]
\centering
\includegraphics[width=\textwidth]{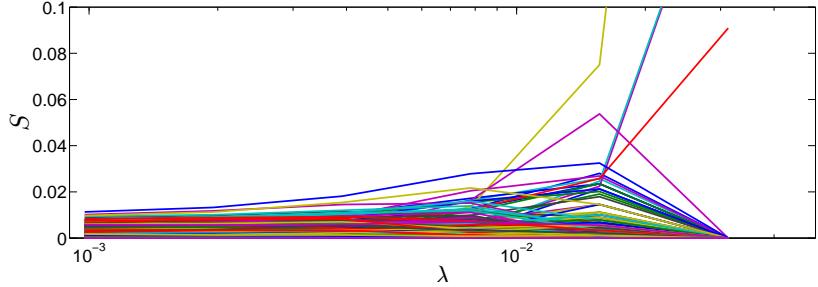}
\caption{$S_g$-scores for all edges, as a function of $\lambda$.}
\label{fig:stab2}
\end{figure}

This randomization-based procedure computes the proability $\Pi_g^{\lambda}$ that an edge $g$ is included in the signature for the parameter $\lambda$. Figure \ref{fig:stab} illustrates these probabilities as a function of $\lambda$ for each edge $g$.
From these probability curves, \citet{meinshausen2008stability} suggests to select the features with the largest maximum probability over $\lambda$.
While this is a nice way to select groups that are robust to the perturbations of the data, we found it hard to apply. Indeed, computation requires to fix a positive lower bound on $\lambda$ and the probability for a given group to exceed the threshold increases when $\lambda$ decreases,  adding an extra parameter to be tuned.
Therefore, we propose a slightly different way to score the groups according to their stability across the perturbations of our data. For each edge $g$ we define the following score:
$$S_g=\max_{\lambda\in\Lambda} \left( \frac{\Pi_g^{\lambda}}{\sum_g\Pi_g^{\lambda}} \right)\,,$$
which is intuitively large for a group that often enters the signature very early, while many others are not yet considered as relevant. Note that this scoring function tends to decrease when $\lambda$ decreases, since more and more groups are selected. Moreover, it constitutes a way not to have to select a value for the regularization parameter. As a matter of fact, figure \ref{fig:stab2} shows the scores that were computed for the groups from figure \ref{fig:stab}. It is clear from this picture that most groups in the final signature are selected for an early $\lambda$.  

Finally, we obtain a ranked list of edges by decreasing score, which allows us to define signatures of various sizes by selecting the groups whose scores are above a threshold. We note that, without stability selection, Lasso and graph Lasso also provide a ranked list of genes, in the order in which they enter the signature.

\subsection{Preprocessing}\label{sec:preproc}
In order to limit the computational burden and discard irrelevant genes we apply the following preprocessing steps each time a signature is built on a training set of gene expression.
\begin{itemize}
\item\textbf{Scaling.} Each gene is scaled to mean zero and variance one.
\item\textbf{Outliers.} For each gene, we remove the outliers from the training set, i.e. for each gene $g$, the examples in set $I$ are removed with $I=\{i, |x_{i,g}|>1.96\}$. We then compute the correlation between the gene expression and the response.
\item\textbf{Threshold.} We keep the $n_g$ genes with the greatest correlation with the response. In practice we fix $n_g=1500$
\item\textbf{Genes kept.} Among the $n_g$ genes, we discard those that are not connected to any other genes in the gene network. This is to ensure that all genes have the possibility to get connected when the signature is built.
\end{itemize}

\subsection{Postprocessing and accuracy computation}
Given a signature $\mathcal{A}$, we estimate a predictive model by fitting a logistic regression. The performance is estimated by $5$-fold cross-validation, in terms of balanced accuracy, i.e. $(sensitivity + specificity)/2$. 

\subsection{Connectivity of a signature}
To quantify whether a set of genes is connected on the gene network, we compute the following connectivity score:
\begin{equation} \label{eq:connectivity}
C_{\mathcal{A}}=\frac{\text{Size of the greatest connected component}}{\text{Number of genes selected}}
\end{equation}
The larger this score, the more connective the solution. The maximum score $1$ is obtained if the active set consists of one and only connected component.

\section{Data}\label{sec:material}

We work on the Van't Veer breast cancer data set from \citet{Vijver2002gene-expression}, and on the Wang dataset from \citet{wang2005gene}, both restricted to $8,141$ genes by \citet{Chuang2007Network-based}. The Van't Veer set contains $295$ tumors, split into $78$ metastatic and $217$ non-metastatic ones, while the Wang dataset contains $286$ tumors among which $106$ are metastatic.

We borrow from \citet{Chuang2007Network-based} a human protein-protein interaction network comprising $57,235$ interactions among $11,203$ proteins, integrated fom yeast two-hybrid experiments, predicted interactions via orthology and co-citation, and curation of the literature.

\OMIT{
For the groups, we consider edges from the latter graph, chromosomic regions and canonical pathways from MsigDB. 
}

\section{Results}

Throughout this section, we use the Lasso as a baseline method for gene selection, and are
interested in the effect of using the graph information and the
stability selection on three main quantities. Our first criterion is the predictive accuracy
obtained by each algorithm. This accuracy is estimated by the
standard $5$-fold \emph{cross-validation} procedure, where the data is
split into $5$ parts, and each part is used to evaluate the
performance of a model which is trained on the union of the $4$
others. We use the same folding in all the experiments, and make sure
that the ratio of metastasic and non-metastasic prognosis is the same
across the $5$ parts. Second, we consider the \emph{stability} of signatures. This involves both the stability within a dataset with respect to random perturbations of the training set, which we estimate by the number of selected genes
that are common to the five folds, and the stability across two
different datasets, which we estimate by comparing the signatures estimated on the Van't Veer and on the Wang datasets.  Finally, we assess how connected the signature is
on the biological graph of~\cite{Chuang2007Network-based}, as an indicator of its interpretability.

\OMIT{
Throughout this section, we use the same splitting for the data,
namely, we divide it into $5$ balanced folds, i.e. each has with the
same number of positive and negative examples.

We are interested in comparing three main quantities: first, the
accuracy obtained with each algorithm. Second, we consider both the
\emph{inner} stability of the signatures, i.e. the number of selected
genes that are common to the five folds, and the stability across two
different datasets.  Finally, since we have the use of a graph, we are
also interested in the connectivity of the selection, that is, the
number and the size of the connected components induced by the
signatures.
}

\subsection{Preprocessing facts} 
\label{sec:preprocfacts}

Before further investigating the results, it is worth noting that
after the preprocessing step where $1500$ genes are kept in each
fold, only $355$ genes (connected through $901$ edges) appear in the five folds after
applying the procedure described in Section~\ref{sec:preproc} on Van't
Veer data.  On the Wang dataset, this reduction is even more
dramatic~: only $145$ genes connected by $97$ edges are selected in all folds. This illustrates
the high instability of the gene selection when changing even partly
the set of patients on which the selection is made. This also
upper-bounds the stability which is obtained by the learning
algorithm, since some genes which are selected on one fold may not be
present in another fold in the first place. Since the
selection made in preprocessing does not follow the same
criterion as the learning algorithm which selects the signature, it is technically possible
that some genes would enter the signature if the preprocessing step was skipped. However, it is quite unlikely that the
instability which is observed on the pre-processing procedure would be
much reduced by directly using the learning algorithm.

Regarding the upcoming assessment of the stability across the
datasets, it is worth pointing out that, after pre-processing, the Van't Veer and Wang datasets have only $118$ genes in common, connected by $78$ edges.

\subsection{Accuracy}

\begin{figure}[!ht]
\centering
\includegraphics[width=\textwidth]{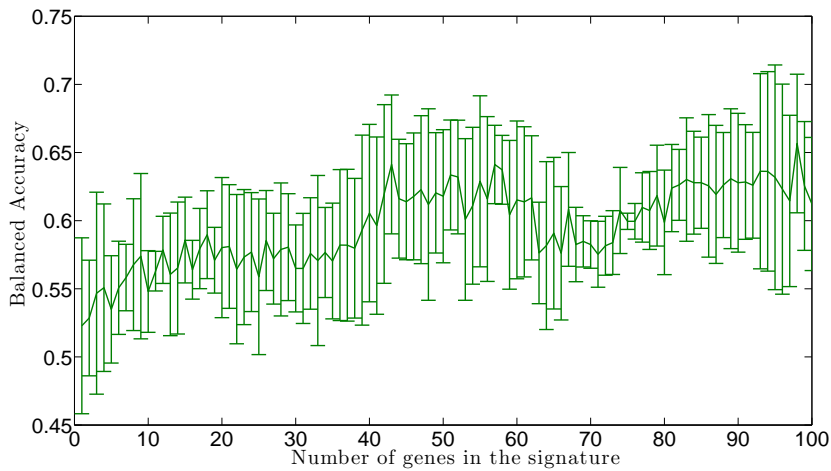}
\caption{Balanced accuracy of the unpenalized logistic regression
  model trained on the signature selected by the Lasso as a function
  of the size of the signature.}
\label{fig:acclasso}
\end{figure}

\begin{figure}[!ht]
\centering
\includegraphics[width=\textwidth]{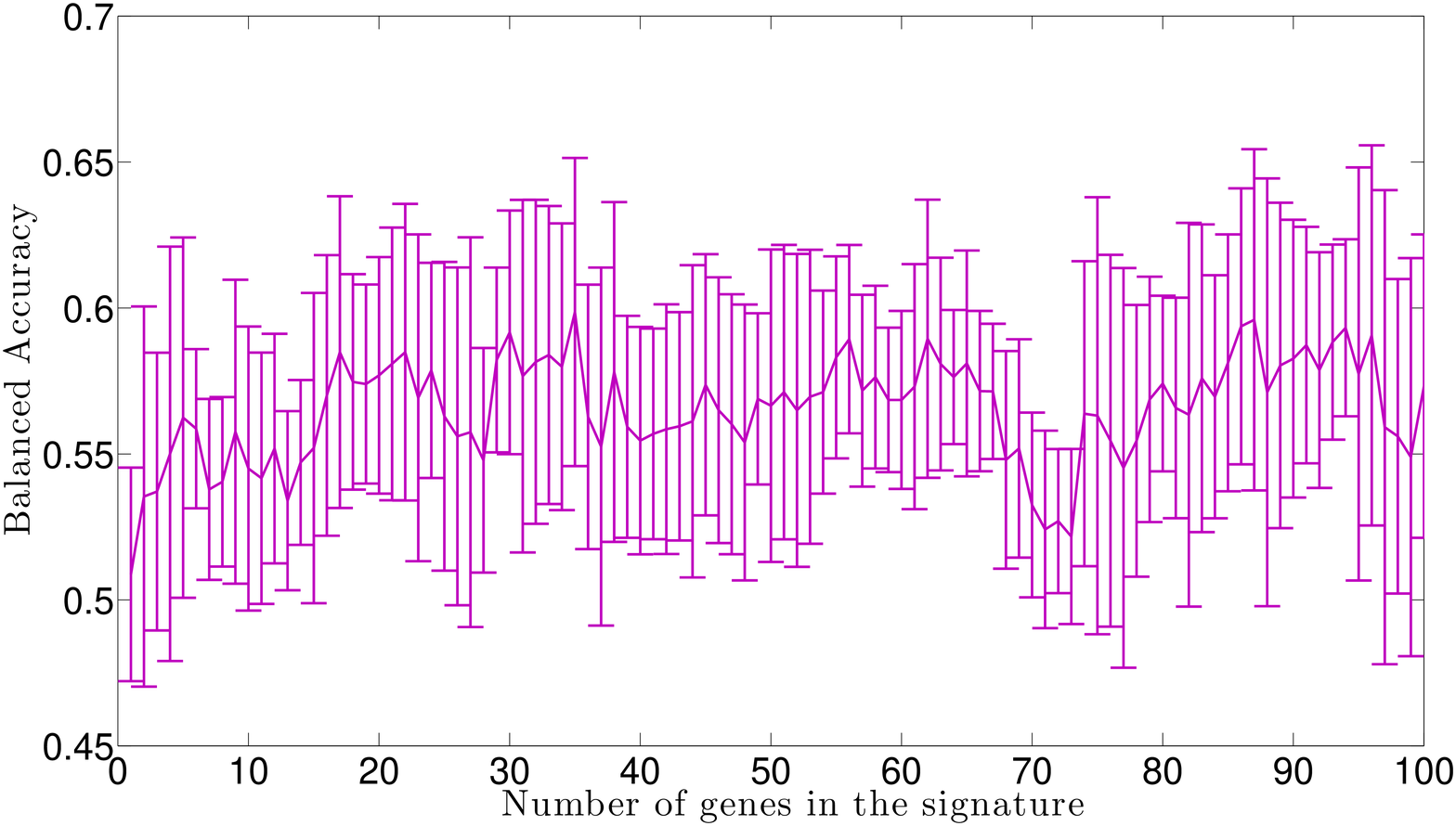}
\caption{Balanced accuracy of the unpenalized logistic regression
  model trained on the signature selected by the Lasso with stability
  selection, as a function of the size of the signature.}
\label{fig:acclassoss}
\end{figure}

\begin{figure}[!ht]
\centering
\includegraphics[width=\textwidth]{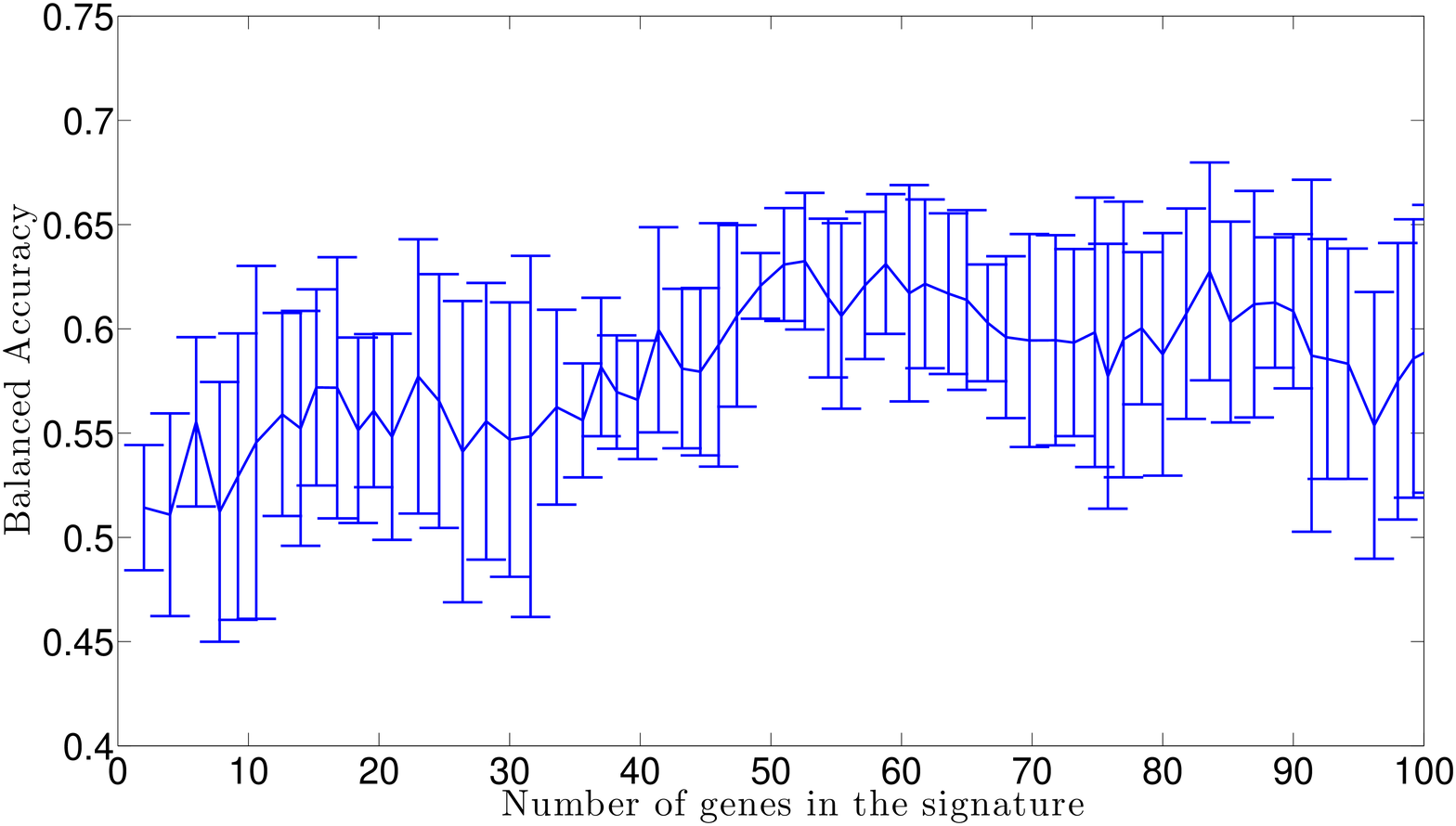}
\caption{Balanced accuracy of the unpenalized logistic regression
  model trained on the signature selected by the graph Lasso as a
  function of the size of the signature.}
\label{fig:accCV}
\end{figure}

\begin{figure}[!ht]
\centering
\includegraphics[width=\textwidth]{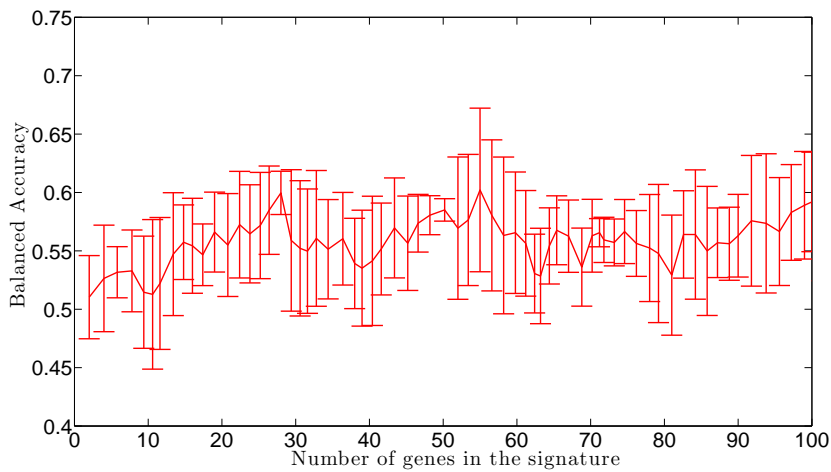}
\caption{Balanced accuracy of the unpenalized logistic regression
  model trained on the signature selected by the graph Lasso with
  stability selection, as a function of the size of the signature.}
\label{fig:accStab}
\end{figure}

Figures~\ref{fig:acclasso},~\ref{fig:acclassoss},~\ref{fig:accCV}
and~\ref{fig:accStab} illustrate the 5-fold cross-validation performances on the Van't Veer dataset for the four gene selection
algorithms, i.e., respectively the Lasso, the Lasso with
stability selection, the graph Lasso and the graph Lasso with
stability selection. We plot the balanced accuracy as a function of the size of the signature. 

All curves look quite similar. For all methods, we observe that the performance degrades when signature is too small. It appears that the accuracies are overall
very similar, \emph{i.e.} neither the use of the graph information
through the graph lasso penalty not the stability selection procedure
significantly change the performance.

In all cases, signatures
with less than $30$ genes are less performant. However, there does not
seem to be a clear number of genes that comes out as the best performer. We
decide to look further into the four signatures of size $60$. It seems a
reasonable size according to the signatures proposed in the
literature.

For each of these four signatures, we now check whether they are also a useful signature on the independent Wang dataset. We thus train
four classifiers on the Wang dataset described in
Section~\ref{sec:material}  restricted to the genes present in each of the four
signatures obtained on Van't Veer dataset. We also train four
classifiers using the same algorithms as the ones used to generate the
signatures on the Wang dataset directly. The objective is to assess
what we lose when selecting the genes on a different dataset for the
four algorithms.

\begin{figure}[!ht]
\centering
\includegraphics[width=\textwidth]{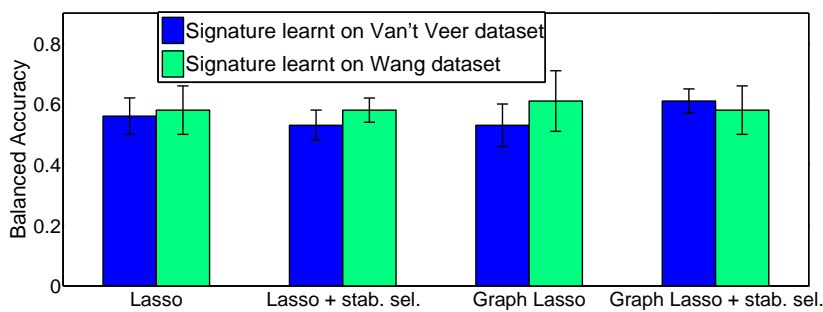}
\caption{Balanced accuracy on the Wang dataset when selecting the
  genes on Wang (green) and Van't Veer (blue) datasets for the four
  algorithms.}
\label{fig:acconwang}
\end{figure}

The results obtained are shown on Figure~\ref{fig:acconwang}. They suggest that signatures estimated on the Van't Veer dataset are in fact almost as good on Wang as signatures estimated on Wang itself, if not better in the case of the graph Lasso with stability selection procedure. 

\subsection{Stability}

Here we compare the stability of gene selection by the algorithms,
i.e. our concerns are both the number of genes selected frequently in
the five folds and the intersection of the signatures learnt on two
different sets of data.

\begin{figure}[!ht]
\centering
\includegraphics[width=\textwidth]{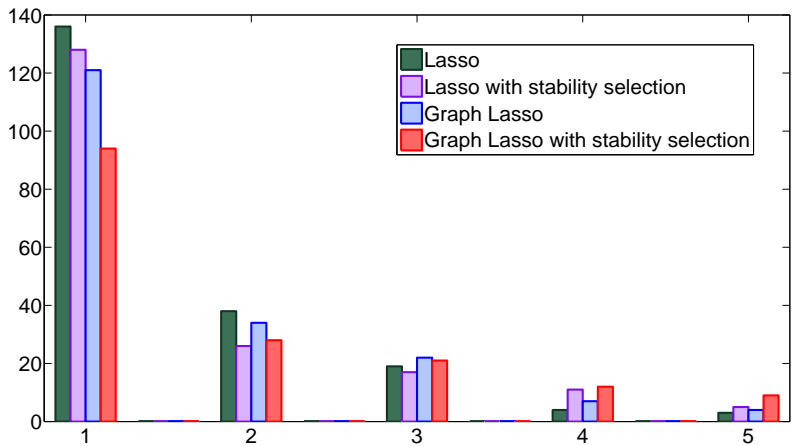}
\caption{Number of genes present in exactly $1,2,3,4$ and $5$ of the
  $5$ folds for the four algorithms.}
\label{fig:hist}
\end{figure}

Figure~\ref{fig:hist} which shows how many genes are in the signatures
of 1, 2, 3, 4 or 5 of the five folds, for each algorithm. A stable
feature selection method should have more genes occurring five times,
and less genes occurring only once.

From a stability point of view, a first improvement over the Lasso is due to the grouping of the variables, as the graph Lasso shows more overlap of more than three folds. However, it appears clearly that stability selection further improve the number of overlaps. Thus, the best stabilization performance is logically obtained by the graph Lasso with stability selection, that combines these two advantages. 

Obviously, even though grouping and randomization give better stability results, the solution is still very inconsistent across folds. We believe that this might be due to the heterogeneity of our dataset, more precisely to the fact that there are different tumor subtypes which we consider altogether instead of as many  as there are subtypes. However the small size of our data set does not allow us to do so. 

\begin{figure}[!ht]
\centering
\includegraphics[width=\textwidth]{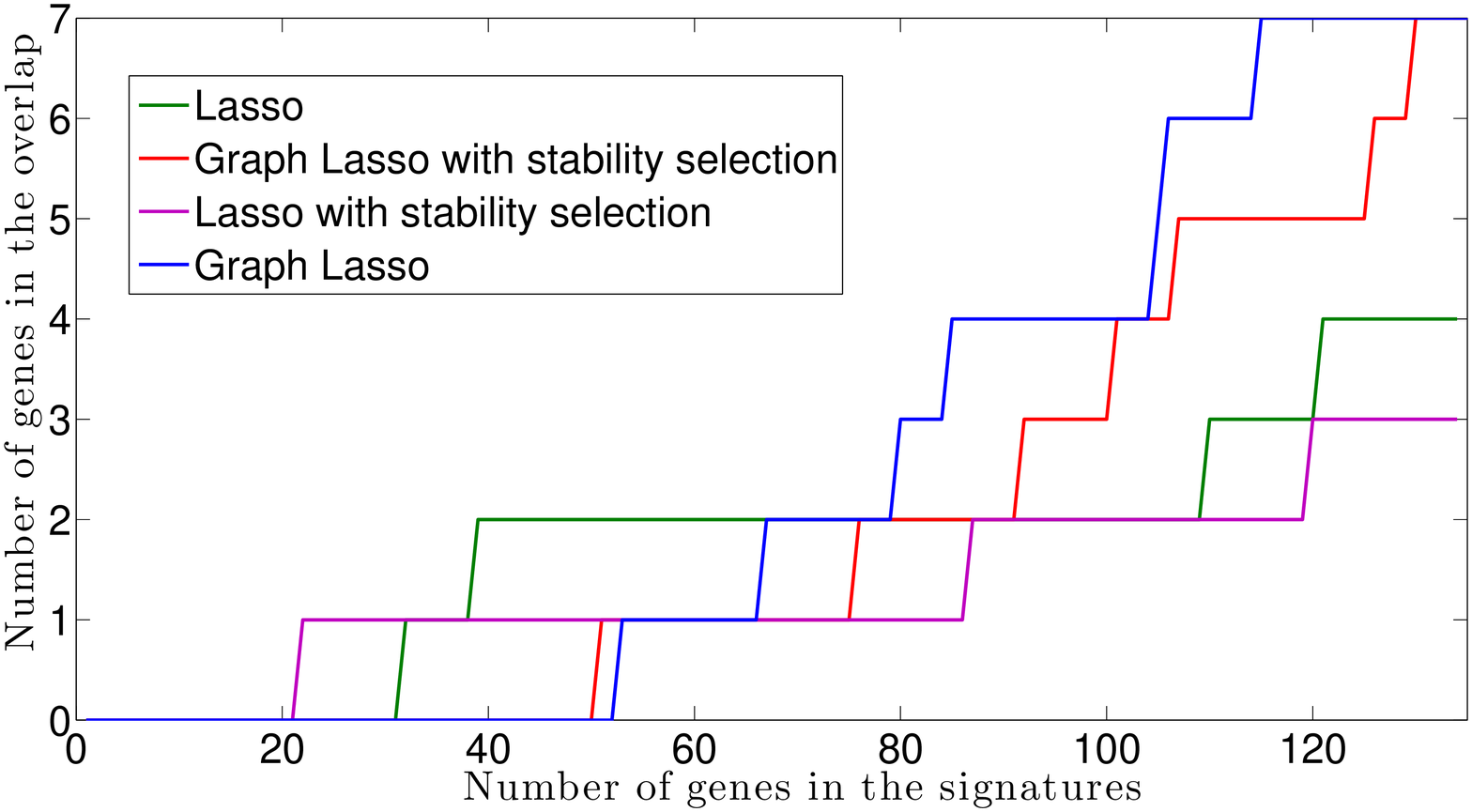}
\caption{Number of genes present in both the signature generated on
  the Van't Veer and the Wang datasets, as a function of the number of
  genes considered in the signature.}
\label{fig:stabVW}
\end{figure}

A different question is whether these algorithms achieve an overlap between two signatures learnt on different datasets, i.e. for what we may hope in terms of reproducibility or exportability of the signatures. Figure \ref{fig:stabVW} sheds some light on this question for it shows the number of genes found in the two signatures from Van't Veer and Wang datasets respectively. While it seems difficult to achieve overlapping with a signature smaller than a few dozen, grouping variables \emph{a priori} still seems to be a way to improve the reproducibility. Randomization does apparently not improve this type of stability. 

However, even when we do find some genes overlapping between the two signatures, there are very few of them. We believe that there could be two main explanations for this fact. First, the distribution of the tumor subtypes may be very different from a dataset to another, leading to very different overall expression patterns. Second the normalization of the data also probably plays a disrupting role for the matter of stability.

\subsection{Connectivity}\label{sec:connectivity}

\begin{figure}[!ht]
\centering
\includegraphics[width=\textwidth]{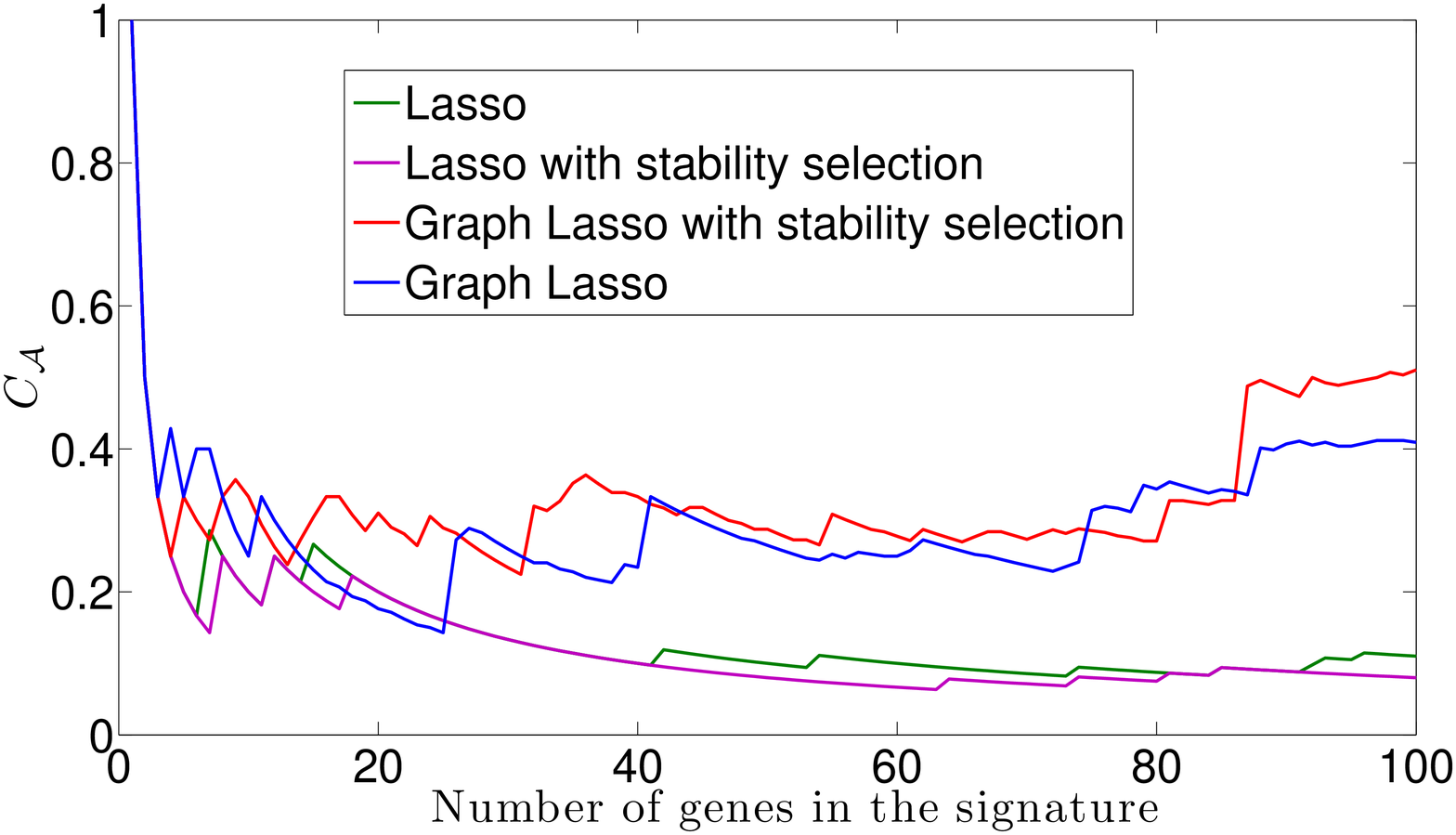}
\caption{Connectivity index of the signatures as a function of the
  number of genes considered in the signature.}
\label{fig:connect}
\end{figure}

Given a graph, it may be interesting to look at the connectivity of the solution, i.e. the number and the size of the connected components induced by a selected signature. Recall that we use the scoring function defined by equation \eqref{eq:connectivity}. First, it is worth noting that both the Lasso ran as a single algorithm and the Lasso with stability selection induce very low connectivity (see figure \ref{fig:connect}). However, it seems that using prior information from a graph, e.g. running either a group Lasso algorithm with edges as groups or that same procedure with stability selection greatly improves the connectivity. Note that using stability selection does not significantly improve the connectivity of the solution. This suggests that mostly the prior is responsible for it i.e. the way to choose the groups, in this case as edges from a graph. 

\begin{figure*}[!ht]
\centering
\includegraphics[width=\textwidth]{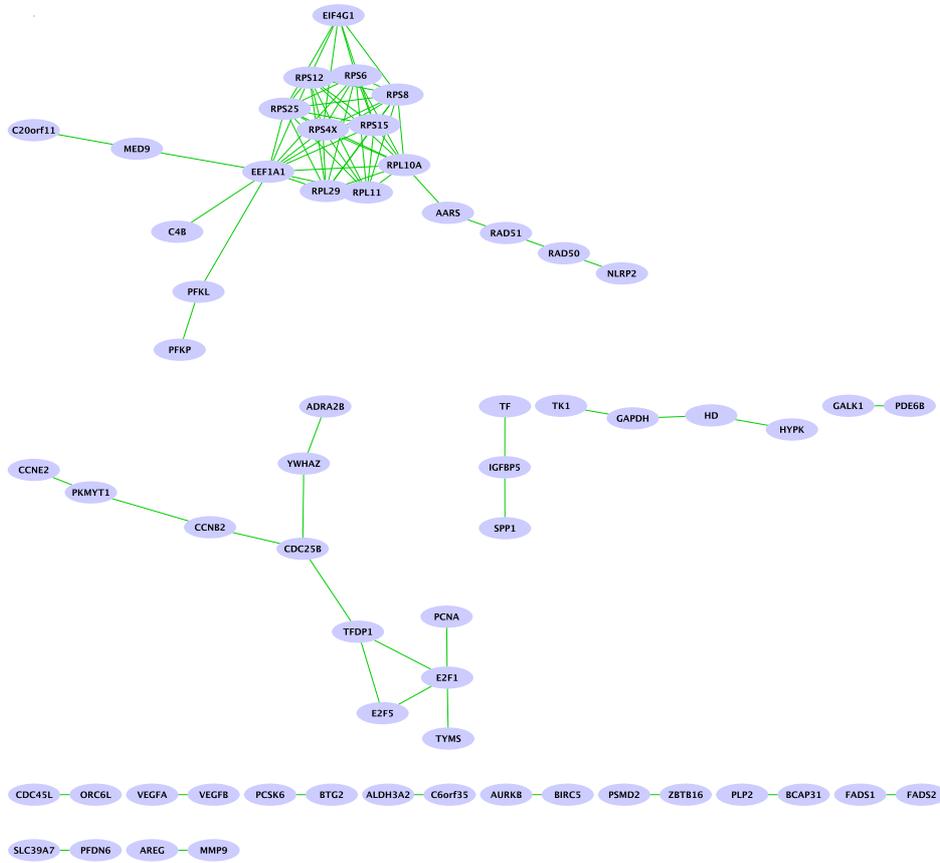}
\caption{Signature obtained with the graph Lasso
algorithm with stability selection.}
\label{fig:signglassoss}
\end{figure*}

Figure~\ref{fig:signglassoss} shows the two $60$-genes signatures
obtained with the graph Lasso with stability selection and the Lasso.

\begin{figure*}[!ht]
\centering
\includegraphics[width=\textwidth]{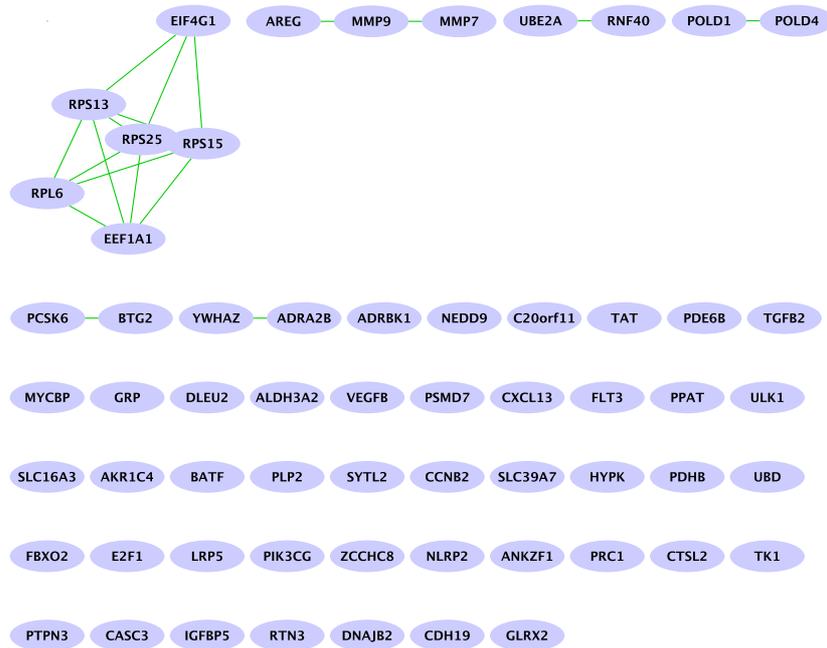}
\caption{Signature obtained with the Lasso
algorithm.}
\label{fig:signlasso}
\end{figure*} 

Obviously, the graph Lasso with stability selection provides a signature that is biologically more relevant than the one chosen with the Lasso. Indeed, the connected components are related to biological processes (see section \ref{sec:bio}) and hence make more sense as a whole. 

\subsection{Biological Interpretation}\label{sec:bio}
Two main connected components are induced by the signature showed in figure \ref{fig:signglassoss}. The largest includes $20$ genes, among which $9$ are ribosomal proteins. This component also includes $RAD50$ and $RAD51$, which are two known DNA repair genes that also belong to the ATMPathway (Tumor Suppressor) and the ATRBRCAPathway along with $BRCA1$ and $BRCA2$. 

The second largest component almost exclusively contains genes involved in cell cycle, such as transcription factor $E2F1$, cyclins $CCNB2$ and $CCNE2$ or cell division cycle gene $CDC25B$.  

Among the $29$ genes left in the signature, two more are involved in cell cyle and five belong to known cancer pathways. 

The second signature (from the Lasso algorithm) is harder to interpret since many genes are singletons. The largest connected component (of size $6$) contains $4$ genes from the ribosomes. $6$ genes in the rest of the signature are known to be involved in some cancer pathways and $4$ belong to the cytokine-cytokine receptor interaction pathway. Overall, the second signature is less interpretable in terms of biological functions than the first one. 

These informations were found using both the KEGG pathways and the canonical pathways from MsigDB.  
\section{Discussion}

In these experiments we assessed the effect of using a biological
graph and stability selection on various characteristics of the
solution. A first important remark is that neither of these methods
significantly improved the estimated prediction accuracy. On the one
hand this is a negative result, as one could have expected that
incorporating prior biological information or selecting more stable
signatures would improve the performance. On the other hand, the
methods are intended to promote the connectivity of the signature on
the graph and making the signature more robust to changes in the set
of patients respectively. Each method seems to succeed at the task it
is intended for~: stability selection tends to produce more stable
signatures accross the $5$ folds and graph lasso outputs signatures
which form a few interpretable connected components on the biological
while signature given by the Lasso essentially gives a list of
disconnected genes which then have to be interpreted independently.
These two improvements are obtained without harming the prediction
accuracy, \emph{i.e.}, these methods allow to obtain signatures which
are as effective as the one output by the Lasso with the additional
benefit of being more stable and more interpretable.

We note however that the obtained signatures remain quite unstable
when changing the set of patients (\emph{e.g.} by considering the
different folds). A first factor which can explain this variability is
the fact that the considered datasets contain several subtypes of
breast cancer tumors, some of which (\emph{e.g.} basal versus luminal) are
considered by practitioners to be distinct diseases, known to involve
distinct biological processes. Finding a unique signature across these
different signals may not be possible, and considering different
models for the different subtypes, or a global model taking these
differences into account may be a better option, although the subtypes
are not strictly defined, and very few patients are available for some
of them.

Another possible explanation is that there does not exist such a small
set of genes which are much more involved than the others in the
process of metastasis, \emph{e.g.} that the underlying signal is not
sparse at the gene level, so that small changes in the dataset give
very different restricted signatures. This of course would not imply
that finding a small set of genes with a good predictive power
(\emph{e.g.} to build prognosis tools) is hopeless, only that there is
no ``true'' signature and that there is no point to looking for something
stable against variations in the dataset. Even in this case, looking
for signatures under some constraints which make them suitable for
analysis, like the one of being connected on a pre-defined graph may
uncover various important aspects of the biological process.



\bibliographystyle{natbib}
%
%

\bibliography{biblio_ismb}

\end{document}